\documentclass[runningheads]{llncs}
\usepackage{graphicx}
\usepackage{url}

\usepackage{algorithm}
\usepackage{fancybox}
\usepackage{fontenc}
\usepackage[most]{tcolorbox}
\usepackage{xspace}
\newcommand{\tool}{\textsc{CHQA}\xspace}
\newcommand{\task}{\textsc{EDC}\xspace}

\makeatletter
\newcommand*{\etc}{%
    \@ifnextchar{.}%
        {etc}%
        {etc.\@\xspace}%
}
\begin{document}
\title{Emotion Detection in Unfix-length-Context Conversation}

\author{Xiaochen Zhang\inst{1*} \and
Daniel Tang\inst{2}}
%
\authorrunning{F. Author et al.}
%
\institute{Australian national university,
\email{xiaochen.zhang@anu.edu.au}\\ \and
University of Luxembourg, China,
\email{realdanieltang.gmail.com}}

\maketitle              

\begin{abstract}
Emotion Detection in conversation is playing a more and more important role in the dialogue system. Existing approaches to Emotion Detection in Conversation (\task) use a fixed context window to recognize speakers’ emotions, which may lead to either scantiness of key context or interference of redundant context. In response, we explore the benefits of variable-length context and propose a more effective approach to \task. In our approach, we leverage different context windows when predicting the emotion of different utterances. New modules are included to realize variable-length context: 1) two speaker-aware units, which explicitly model inner- and inter-speaker dependencies to form distilled conversational context, and 2) a top-k normalization layer, which determines the most proper context windows from the conversational context to predict emotion. Experiments and ablation studies show that our approach outperforms several strong baselines on three public datasets.

\end{abstract}

\section{Introduction}
\begin{figure*}[ht]
  \centerline{\includegraphics[width=0.95\textwidth]{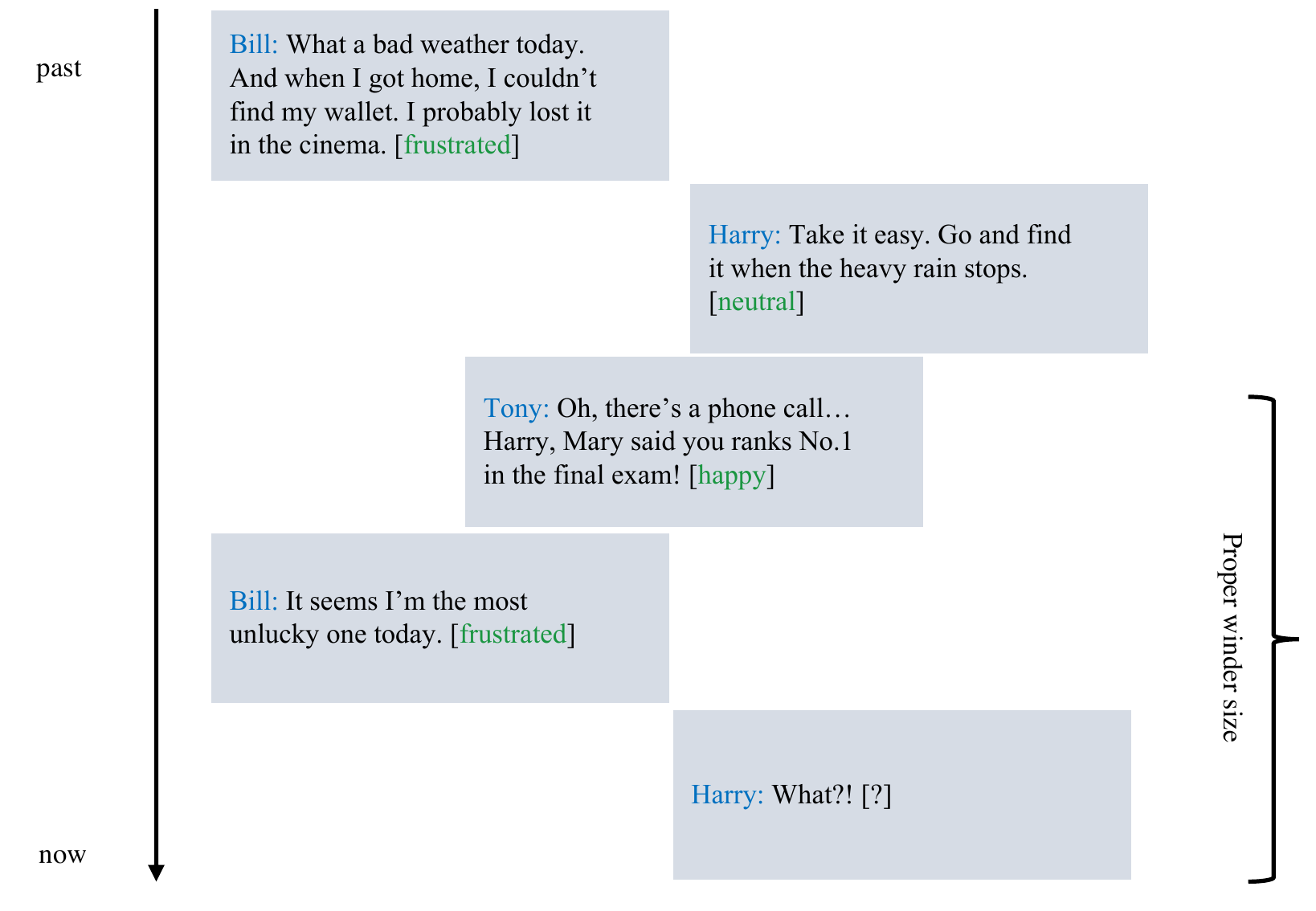}}
  \caption{A multi-party \task example. The ideal context window to Harry’s emotion
would include exactly two preceding utterances, among which Tony provides evidence
for Harry being happy. Utterances ahead of Tony are redundant since they are irrelevant to the current turn of conversation.}
  \label{fig:overview}
\end{figure*}
Emotion Detection in Conversation (\task) is the task of predicting the speaker's emotion in conversation according to the previous context and current utterance. Great technical breakthroughs of \task promote the development of applications in an army of domains, such as healthcare, political elections, consumer
products and financial services \cite{wang2022hienet,tang2021moto,wang2022automatically,hietang,tang2021ckg,liu2012sentiment,nasukawa2003sentiment,poria2019emotion}.
Figure 1 shows an example of \task. Existing approaches \cite{ghosal2020cosmic,ghosal2019dialoguegcn} consider a fixed context window (i.e., the number of preceding utterances), which
may suffer from two issues: (1) semantic missing due to a small window; or (2) redundancy problem in big context text, making it difficult to choose the right context in the task \tool. Therefore, knowing the current speaker is Harry is beneficial to choosing the right context window since one of the preceding utterances explicitly mentions Harry, which indicates that it may contain information relevant to the current utterance. That is, speaker dependencies are the key indicators to determine the right context window. speaker dependencies are both critical to conversation understanding \cite{ghosal2020utterance}, where
speaker dependencies can be further categorized into inner- and inter-speaker dependencies [8]. Firstly, we model the above dependencies in an attention-based
utterance encoder and two speaker-aware units to generate conversational context representation, where inner- and inter-speaker dependencies are explicitly
modeled to help detect the ideal context windows. Next, a top-k normalization layer generates top-k best context windows and their probability weights
based on the dimension-reducted context representation. Lastly, we predict the
emotion of the current utterance by softly leveraging the top-k best context windows. Experiments show that our approach achieves competitive performance
on three public conversational datasets: 66.35\% F1 on IEMOCAP \cite{busso2008iemocap}; 61.22\%
F1 on DailyDialog \cite{li2017dailydialog}; and 38.93\% F1 on EmoryNLP \cite{zahiri2018emotion}. Extensive ablation studies demonstrate the contribution of each component in our approach as well
as the necessity of using variable-length context. 

We summarize our contributions as threefold:
\begin{itemize}
    \item For the first time, we alleviate the context scantiness and context redundancy
problems in \task by varying the length of context.
    \item We propose a new approach that considers different context windows for
different instances to conduct emotion prediction, where 1) speaker dependency is explicitly modeled by new speaker-aware units to
help the detection of ideal context windows and 2) a new top-k normalization
layer that generates top-k best context windows as well as their weights.
    \item We achieve competitive results on three public \task datasets and conduct
an elaborate ablation study to verify the effectiveness of our approach.
\end{itemize}

\section{Related Work}
Recent \task studies are based on Deep Learning, which can be further categorized into three main kinds: RNN-based, GCN-based and Transformer-based
models. RNN-based models have been well explored in the last few years. Poria
et al. (2017) \cite{poria2017context} first modeled the conversational context of \task using Recurrent Neural Networks (RNNs) \cite{medsker2001recurrent}. Hazarika et al. (2018) \cite{hazarika2018conversational} took speaker information into account and Hazarika et al. (2018) \cite{hazarika2018icon} first modeled Inter-speaker
dependencies. Majumder et al. (2019) \cite{majumder2019dialoguernn} kept track of speakers’ states and their
method could be extended to multi-party conversations. Lu et al. (2020) \cite{lu2020iterative} proposed an RNN-based iterative emotion interaction network to explicitly model
the emotion interaction between utterances. Ghosal et al. (2019) \cite{ghosal2019dialoguegcn} and Sheng
et al. (2020) \cite{sheng2020summarize} adopted relational Graph Convolutional Networks (GCN) 
to model \task, where the whole conversation was considered as a directed graph
and they employed graph convolutional operation to capture the dependencies
between vertices (utterances). However, converting conversations to graphs loses
the temporal attributes of the original conversation. Owing to the excellent representation power of transformers \cite{devlin2018bert}, some researchers adapted them to \task and got
favorable results \cite{li2020hitrans}. Recently, Ghosal et al. (2020) \cite{ghosal2020cosmic} incorporated commonsense knowledge extracted from pre-trained commonsense transformers COMET
\cite{bosselut2019comet} into RNNs and obtained favorable results on four public \task datasets. However, none of the above models regarded the context scantiness or the context
redundancy problem as us.

\section{Our Method}

\subsection{Problem Formulation}
A conversation consists of n temporally ordered utterances $\{x_1,\dots,x_n\}$ and their speakers $\{ s_1,\dots,s_n \}$. $x_i$  is the {\it i-th} word in the sequence. At time step {\it t},  
the goal of \task is to identify the most-likely categorical emotion label $\hat{y}_t$ for speaker $s_t$ given the current and preceding utterances as well as their speakers: $\hat{y}_t = argmax p(y_t|x_{1:t}, s_{1:t})$, where $1:t$ means set of the former {\it t} elements.
\begin{figure*}[ht]
  \centerline{\includegraphics[width=0.95\textwidth]{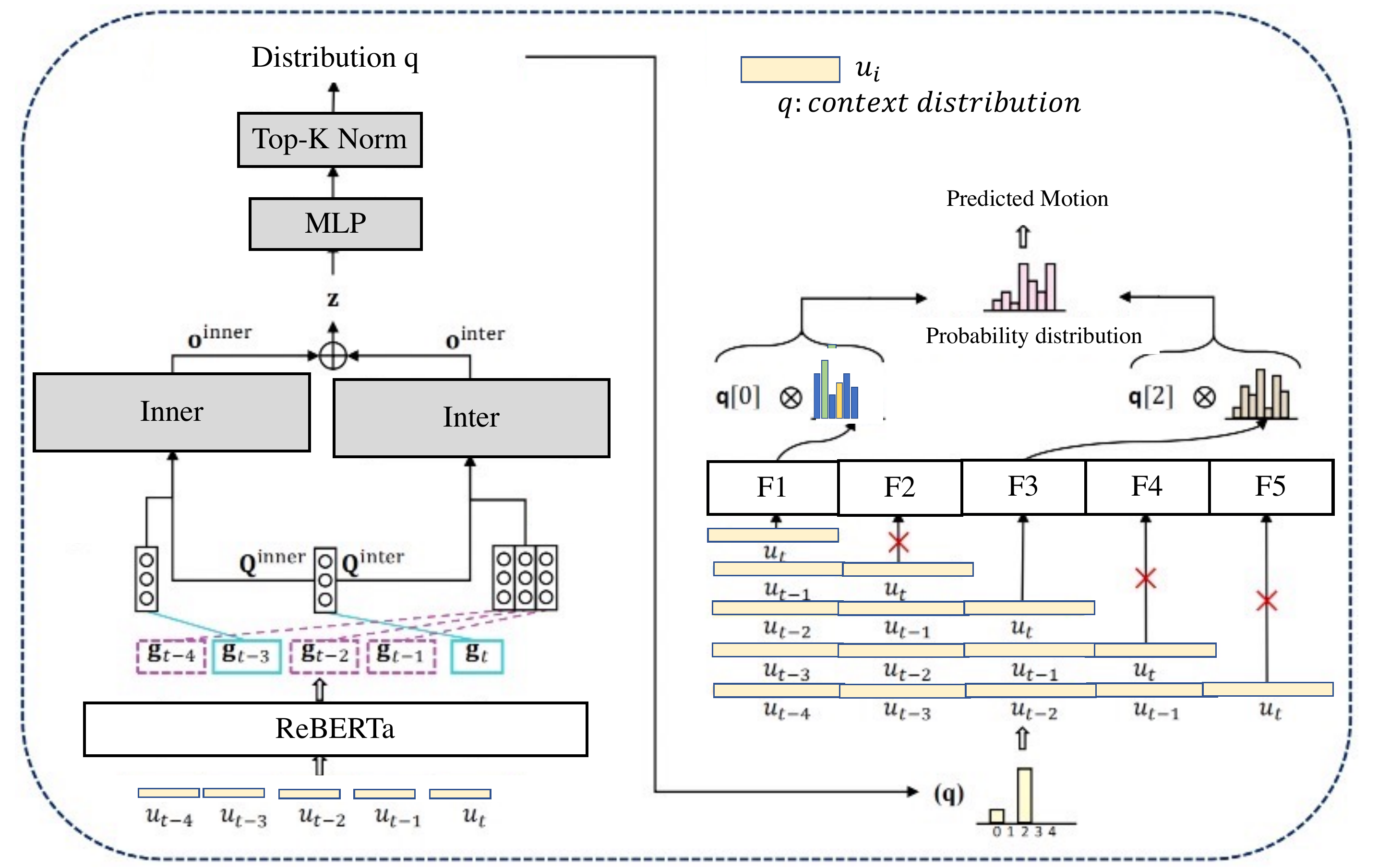}}
  \caption{Overall architecture of our approach.}
  \label{fig:arc}
\end{figure*}
\subsection{Model}
As depicted in Figure \ref{fig:arc}, our approach consists of the following modules: (1) an
utterance encoder that encodes sequential dependencies among utterances; (2)
two speaker-aware units that explicitly encode inner-and inter-speaker dependencies to help detect the ideal context windows; (3) a multi-layer perception
and a top-k normalization layer that generates distribution over different context windows, from which we determine top-k best context windows and their
corresponding weights; and (4) a prediction module that generates emotion distribution from the top-k best context windows with different probability weights.
Utterance Encoder The input of the utterance encoder is a sequence of tokens
with speaker information. At time step t, we generate the input sequence by
prepending speaker information (i.e. the name of the speaker) to each utterance
and then concatenating utterances up to time step t into a single sequence of
tokens. The name of the speaker and the utterance are separated by special [SEP]
token. The input sequence is fed into the base version of Roberta [12] to
encode the sequential dependencies among utterances and generate contextual
representation for each utterance:
\begin{equation}
    \begin{split}
        u_i &= s_i \oplus [SEP] \oplus x_i, \\
        [g_1, \dots, g_t] &= RoBERTa(\oplus^t_{i=1}u_i)
    \end{split}
\end{equation}
where $g_i$ represents the contextual representation for utterance at {\it i}, which is the Roberta output corresponding to the first token of $u_i$ With a
context window considering up to M previous time steps, the encoder outputs
a sequences of vectors $[g_{t-M}, \dots, g_{t-1}, g_t]$, where $g_i \in \mathcal{R}^d$.

Speaker-Aware Units: Our approach incorporates speaker dependencies to
guide the detection of ideal context windows. Concretely, we propose two speaker-aware units to explicitly capture inner-speaker and inter-speaker dependencies.
The two units have the same attention-based structure, but they do not share
parameters. We first divide utterance contextual representations $[g_{t-M}, \dots, g_{t-1}]$, 
into two subsets Ginner and Ginter depending on whether their corresponding
speakers are the same as the current one. Each speaker-aware unit then takes the corresponding subset G and gt as input, and applies multi-head attention
with layer normalization to incorporate speaker dependencies:
\begin{equation}
    \begin{split}
        o &= LayerNorm(c+g_t), \\
        c &= Concat((head_1,\dots, head_h), \Phi_1), \\
        head_i &= Attention((g_t, G, G)^T(\Phi_2, \Phi_3, \Phi_4)), \Phi \in \mathcal{R}
    \end{split}
\end{equation}
where $\Phi$s are the parameters of the different layer in our model. Finally, we concatenate $o^{inter}$ and $o^{inner}$ into the vector $z$ as $z= [o^{inter};o^{inner}] \in \mathcal{R}^2d$.

Context Window Distribution: Using the distilled vector z, we generate a
the probability distribution over context windows ranging from 0 to M. This is
done via: (1) a multi-layer perceptron (MLP) which maps the distilled vector to
scores of context windows, and (2) a top-k normalization layer that generates
distribution over context windows.

Specifically, we first feed the distilled vector z into a two-layer MLP to get
scores of context windows s:

\begin{equation}
    \begin{split}
        h&= ReLU(z, \Phi_5) \in \mathcal{R}^d_h, \\
        s &= MLP(h;\Phi_6) \in \mathcal{R}^{M+1}
    \end{split}
\end{equation}.

Emotion Prediction from top-K best Context Windows Instead of using
the context window with the highest probability to predict emotion, we use $q = softmax(s+m)$
as soft labels and leverage all top-K context windows in prediction.
As shown in Figure 2, our prediction module contains M + 1 context fields from 0 to M, where field i corresponds to the use of context window i. The input of each field, with a \texttt{[CLS]} at its front, is encoded by a field-specific contextual encoder, which has the same architecture as our utterance encoder. We use a field-specific linear classifier to the encoder output for \texttt{[CLS]}, 
\begin{equation}
    g^i_{CLS} \in \mathcal{R}^d
\end{equation}, to compute the emotion label distribution $p^i$ given context window i:

\begin{equation}
    p^i = softmax(g^i_{[CLS]}; \Phi_7) \in \mathcal{R}^c.
\end{equation}

The final emotion label  distribution $\hat{p}$ combines top-K context window distribution and emotion label distributions given different context windows:
\begin{equation}
    \hat{p} = \Sigma_{i\in top-K} 1[i]p^i \in \mathcal{R}^c.
\end{equation}

\subsection{Training}
We optimize cross-entropy loss  \begin{math}\mathcal{L}\end{math} for each mini-batch \textbf{B} of conversations:
\begin{equation}
    \mathcal{L} = \Sigma^{|B|}_{i=1}\Sigma^{|B_i|}_{j=1} -log \hat{p}^{ij}[y_{ij}],
\end{equation}

\section{Experiment Design}
\subsection{Dataset}
We evaluate our approach on four publicly available datasets, IEMOCAP \cite{busso2008iemocap},
DailyDialog \cite{li2017dailydialog}, MELD \cite{poria2018meld} and EmoryNLP \cite{zahiri2018emotion}. They differ in the number of interlocutors, conversation scenes, and the emotion labels. As shown in
Table 1, the average conversation lengths of the four datasets differ a lot, with the
maximum 49.23 for IEMOCAP and minimum 7.85 for DailyDialog. Moreover,
the datasets hold varied data capacity and average utterance lengths. Following existing approaches, models for the datasets are independently trained and
evaluated. For preprocessing, we follow Zhong et al. (2019) \cite{zhong2019knowledge} to lowercase and tokenize the utterances in the datasets using Spacy.

\subsection{Baselines}
To demonstrate the effectiveness of our approach, we compare it with several
strong baselines as follows:

\begin{itemize}
    \item DialogueRNN, an RNN-based ERC model that keeps track of the states of context, emotion, speakers, and listeners by several separate GRUs.
    \item DialogueGCN, a GCN-based ERC model, where adopt relational graph neural networks to model different types of relations between utterances in the conversation according to their temporal order and speakers. 
    \item KET, a transformer-based model, which leverages external knowledge from emotion lexicon NRC VAD  and knowledge base ConceptNet to enhance the word embeddings. They adopt a hierarchical attention-based strategy to capture contextual information.
    \item RoBERTa-BASE, the base version of Roberta. The inputs are concatenated utterances and the representation of the first subword from the last layer is fed to a simple linear emotion
    classifier. If the input length exceeds the limitation of Roberta, we discard the remote utterances at the utterance level.
    \item COSMIC, a strong ERC model which extracts relational commonsense features from COMET  and utilizes several GRUs to incorporate the features to help emotion classification.
\end{itemize}

\section{Experimental Result}

\subsection{main results}
\begin{figure*}[ht]
  \centerline{\includegraphics[width=0.95\textwidth]{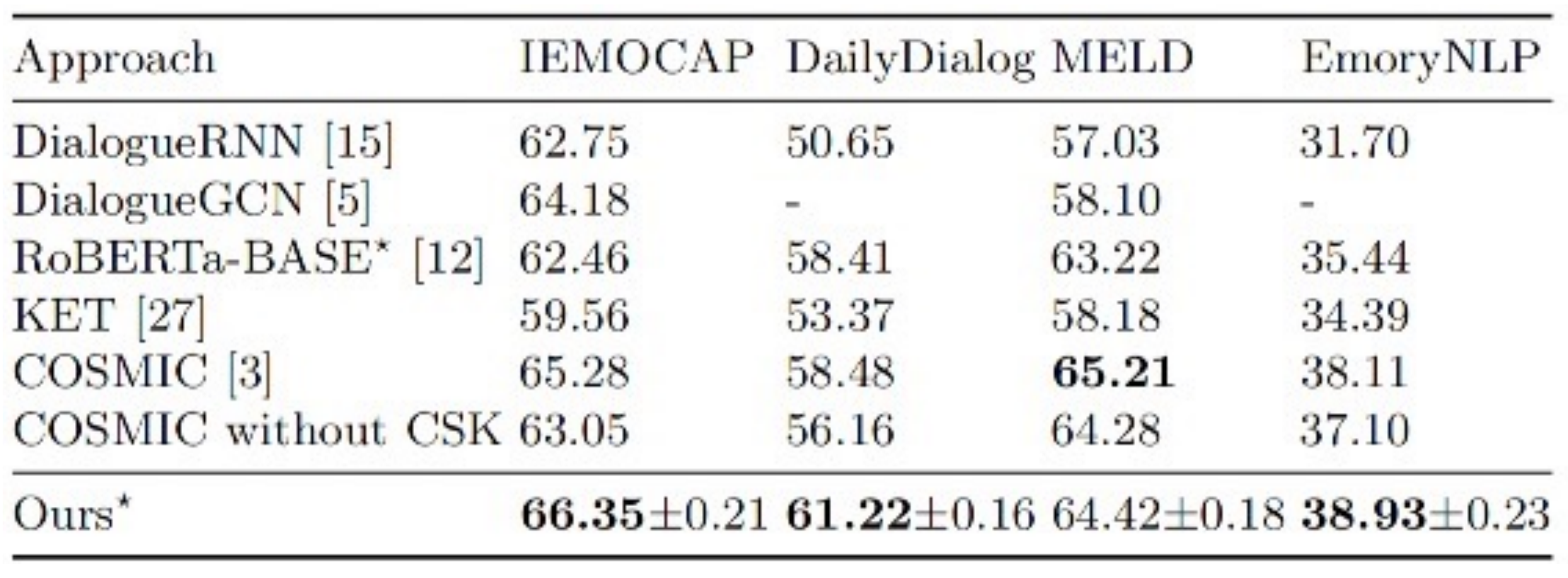}}
  \caption{Main results. The best F1 scores are highlighted in bold. - signifies the unreported results. CSK is the abbreviation of commonsense knowledge. $\star$ means the results obtained by our implementation}

  \label{fig:tb1}
\end{figure*}

The main results are reported in Table 2. Our approach achieves the best performance on IEMOCAP, DailyDialog and EmoryNLP datasets, surpassing COSMIC by 1.07\%, 2.74\%, and 0.82\% F1 scores respectively. We owe the better performance of our approach over COSMIC to the consideration of variable-length context. Moreover, unlike COSMIC, our approach does not rely on external knowledge. For MELD, the result of our approach is also competitive, outperforming all
the baselines except COSMIC. We show that the slightly better performance
of COSMIC is due to the use of commonsense knowledge (CSK). Our approach
performs better than COSMIC without CSK. This indicates that external knowledge could be beneficial to the prediction of short utterances. We leave adding

\subsubsection{Ablation Study}

\begin{figure*}[ht]
  \centerline{\includegraphics[width=0.95\textwidth]{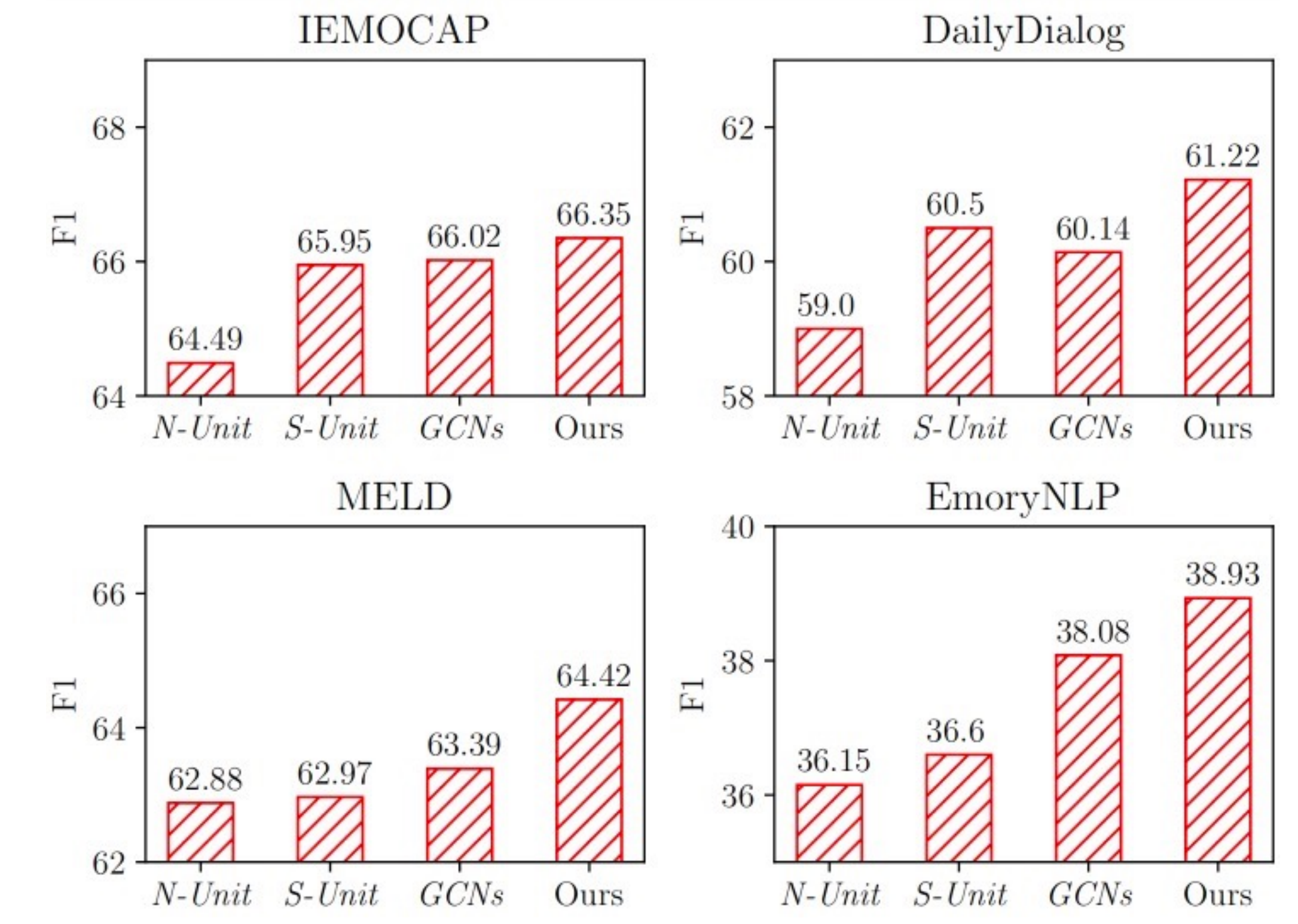}}
  \caption{Ablation for the speaker-aware units on the test sets of four datasets.}
  \label{fig:ablation}
\end{figure*}

In order to expose the contribution of different components in our approach, we
conduct ablation experiments on the main components: the speaker-aware units
and the generation method of context window distribution.
Speaker-Aware Units We compare the speaker-aware units with the following
modeling methods of speaker dependencies:
N-Unit: N-Unit shares the same structure with the inner- (inter-) speaker-aware unit. Different from the speaker-aware units, the keys, and values of its
inputs are all the previous utterance representations regardless of their inner and inter-speaker relationships. N-unit is non-speaker-aware.
S-Unit: S-Unit concatenates one-hot vectors, which indicates the speaker of
each utterance, to the utterance representations and conducts the same operation
as N-Unit.
GCNs: Method from [5], where multiple graph convolution layers capture the
speaker dependencies. Nodes are utterances and edge weights are obtained by
a similarity-based attention module. We add a max pooling layer and a linear
layer after it to get the vector z. The inputs of GCNs are the outputs of our
utterance encoder.
Fig. {fig: ablation} shows the comparison results. We attribute the superior performance
of our method over S-Unit to the explicit modeling of inner- and inter-speaker
dependencies. S-Unit surpasses N-Unit, indicating that speaker information is
indispensable in the context modeling of \task. Moreover, our speaker-aware
units gain over the best of the other three methods by 0.33\% and 0.72\% F1
scores on dyadic datasets (IEMOCAP and DailyDialog), less than those on multiparty datasets (MELD and EmoryNLP), 1.03\% and 0.85\%. We attribute this to more complex speaker dependencies in multi-party conversations than dyadic
conversations. Our method is better at capturing speaker dependencies when
more speakers participated in the conversation.
The generation method of Context Window Distribution q (see Equation 11) controls the activation of context fields and acts
as attention, weights to merge the output distributions of activated context fields.
In our method, we adopt an MLP and a top-k normalization layer to generate
q. We try several other generation methods of q and compare them with our
method. Based on the two functions of q, top-k activation of context fields, and
output distribution weighting, we consider the following variants of our method:

\noindent
All-Soft: The top-k normalization layer in our method is replaced by a softmax layer to get q, which means that all of the M + 1 context fields are always activated and the output distributions of context fields are merged by attention weights.

\noindent
Top-Hard: After the top-k normalization layer, K non-zero probabilities in q are
set to 1K, meaning that the output distributions of K-activated context fields
are weighted equally.

\noindent
All-Hard: Regardless of the sequential and speaker dependencies, all the probabilities in q are set as M1+1, which means that all of the M + 1 context fields are always activated and the output distributions of context fields are weighted equally.

\noindent
Topk-Soft: Method in our proposed approach. F1 scores of the test sets are shown in Fig {fig:tb2}. Compared to All-Hard, AllSoft only has better performance on EmoryNLP. We attribute this to the fact
that the attention weights of proper context windows are not significantly larger
than those of improper ones. Therefore, directly deactivating improper context fields in our approach is more reasonable than activating them and giving them
less attention weights. In response to the above analysis, Topk-Hard outperforms
All-Hard nearly across all the datasets, indicating again that we should avoid
activating improper context fields. Our top-k normalization layer promotes the
attention weights of the K-activated context fields, which is signified by the superior performance of Topk-Soft over Topk-Hard.
According to the above analysis, our generation method of context window
distribution not only avoids activating improper context fields but also gives the
activated ones more reasonable attention weights. As a result, our method outperforms other generation methods. How to further reduce the attention weights
of improper context fields deserves more exploration in the future.

\begin{figure*}[ht]
  \centerline{\includegraphics[width=0.95\textwidth]{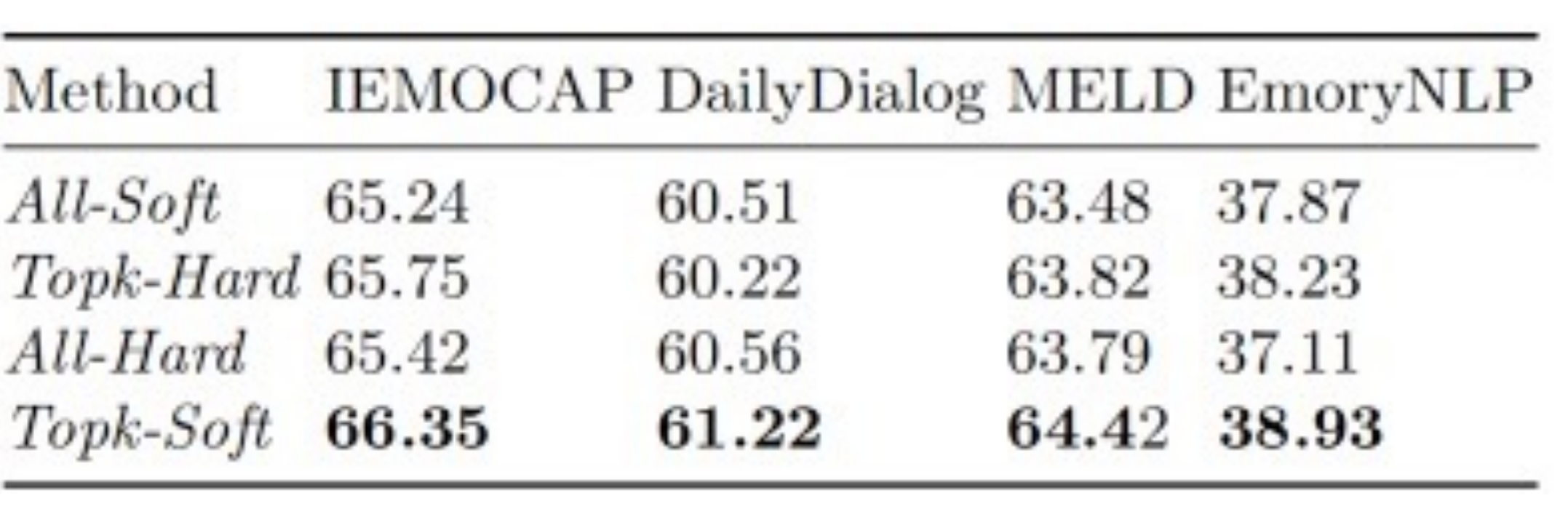}}
  \caption{Ablation for the generation method of context window distribution on the
test sets of four datasets.}
  \label{fig:tb2}
\end{figure*}
\section{Conclusion}
To alleviate the context scantiness and context redundancy problems in \task,
we present a new \task approach capable of recognizing speakers’ emotions
from variable-length context. In our approach, we first generate a probability distribution over context windows according to sequential and speaker dependencies, where speaker dependencies are explicitly modeled by the newly proposed
inner- and inter-speaker units. Then, we introduce a new top-k normalization
layer to leverage all top-k best context windows to conduct emotion prediction
conditioned on the context window distribution. Elaborate experiments and ablation study demonstrate that our approach can effectively alleviate the context
scantiness and context redundancy problems in \task while achieving competitive performance on three public datasets. In the future, we tend to improve the
context window distribution through external knowledge or auxiliary tasks. Also, we’ll
explore more effective mechanisms for the detection of proper context windows.

\bibliographystyle{splncs04}
\bibliography{ref.bib}

\end{document}